\documentclass[lettersize,journal]{IEEEtran}
\usepackage{amsmath,amsfonts}
\usepackage{algorithmic}
\usepackage{algorithm}
\usepackage{array}
\usepackage[caption=false,font=normalsize,labelfont=sf,textfont=sf]{subfig}
\usepackage{textcomp}
\usepackage{stfloats}
\usepackage{url}
\usepackage{verbatim}
\usepackage{multirow}
\usepackage{hyperref}
\usepackage{graphicx}
\usepackage{textcomp}
\usepackage{xcolor}
\usepackage{cite}
\begin{document}

\title{FlexQuant: Elastic Quantization Framework for Locally Hosted LLM on Edge Devices}

\author{
Yuji Chai*, Mujin Kwun*, David Brooks, Gu-Yeon Wei \\
\textit{Harvard John A. Paulson School Of Engineering And Applied Sciences} \\
Cambridge, Massachusetts, USA \\
yuc927@g.harvard.edu, mujin\_kwun@harvard.edu

}

\maketitle

\begin{abstract}
Deploying LLMs on edge devices presents serious technical challenges. Memory elasticity is crucial for edge devices with unified memory, where memory is shared and fluctuates dynamically. Existing solutions suffer from either poor transition granularity or high storage costs. We propose FlexQuant, a novel elasticity framework that generates an ensemble of quantized models, providing an elastic hosting solution with 31x more deployment options, 15x granularity improvement, and 10x storage reduction compared to SoTA methods. FlexQuant works with most quantization methods and creates a family of trade-off options under various storage limits through our pruning method. It brings great performance and flexibility to the edge deployment of LLMs.
\end{abstract}


\section{Introduction}
\label{section-introduction}

Since the introduction of ChatGPT, countless companies in the technology sector have been releasing personal AI assistant products. While OpenAI hosts ChatGPT purely in the cloud, many of these companies are transitioning toward hosting AI services locally on edge devices to deliver truly personalized AI experiences.

However, when deploying LLMs locally, managing memory footprint is one of the main bottlenecks: LLMs are hungry for memory usage, while edge devices are usually limited in on-device memory resources. 

This mismatch has made model quantization the default for edge deployment.
However, reducing the memory footprint through quantization merely solves half of the problem; the other half is memory footprint elasticity.

LLMs are routinely deployed on a diverse range of hardware devices with varying memory constraints. Total system memory can range from 32GB on a laptop to as little as 1GB on embedded devices.

To operate across this range, LLMs must adapt their footprint to the device, elastically scaling model size to accomodate available resources.

In addition to deployment elasticity, locally hosted LLMs must remain elastic to dynamically adapt to the runtime environment.
Most edge SoCs utilize unified memory architectures \cite{orin, jetson, apple}, so the memory available at the time of a request may be uncertain and will fluctuate depending on competing memory usage from other applications. 
Thus, memory elasticity during runtime is another necessity. 
In this work, we focus our efforts on providing these two types of elasticity with one coherent framework. 

\begin{figure}[t!]
\includegraphics[width=0.48\textwidth]{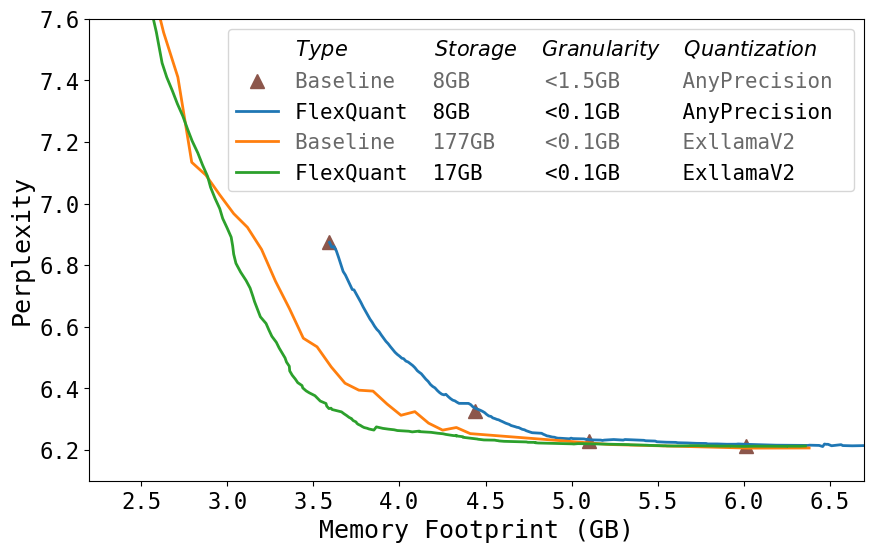}
\vspace{-1em}
\caption{Perplexity comparison of different elastic hosting methods of quantized Llama $2$ $7$B model. It compares FlexQuant with the baseline method when they are using two quantization methods, ExLlamaV2 and AnyPrecision.}
\label{figure:intro}
\end{figure}
\IEEEpubidadjcol 

To enable such elasticity, researchers have designed many compression methods, such as GPTQ \cite{frantar22}, \href{https://github.com/turboderp/exllamav2}{ExllamaV2}, SmoothQuant\cite{xiao23}, etc. These methods, especially ExllamaV2, offer system engineers a flexible design tool that can be used to generate an ensemble of quantized LLMs with various memory footprints. 
Switching between different models in that ensemble serves as an elastic hosting solution, but this baseline method requires 100GB+ of storage to store all the models.
A typical edge device, such as a cell phone, a tablet, or a laptop, usually only has a few hundred GB of storage, so a 100GB+ footprint is too costly. 
A recent work, AnyPrecision\cite{park24}, provided a better solution in which different models could share their parameters, resulting in an elastic policy requiring less than 10GB of storage. However, it suffers from limited LLM deployment options with poor granularity and a worse memory-perplexity trade-off compared to ExLlamaV2, as demonstrated in \autoref{figure:intro}. The largest memory footprint gap between two options is larger than 1.5GB. 
As shown in our analysis in \hyperref[section:granularity]{Section II-C}, that gap is far from sufficient. 
Based on those observations, we propose our elasticity framework, FlexQuant, with the following contributions: 
\begin{enumerate}
\item A 15x improvement in transition granularity and 31x more memory footprint options, with the same storage cost, compared to the SoTA elastic hosting methods, as shown in \autoref{figure:intro}.
\item A search algorithm that further improves the memory-accuracy trade-off pareto frontier, while achieving a storage cost reduction of more than 10x and retaining the same level of granularity, as shown in \autoref{figure:intro}.
\item A pruning strategy that maintains LLM performance while further reducing the total storage cost of the elastic policy by 40\%. 
FlexQuant generates a family of elastic hosting options with different storage costs, which provides more flexibility for system designers.
\end{enumerate}

\section{Background}
\label{section-background}

\subsection{LLM on Edge Devices}
Major industry players are launching their locally hosted LLMs on their personal computing platforms. 
Apple has integrated Apple Intelligence, their AI assistant service, on their entire hardware product line \cite{apple24}. 
Google has released their Google AI Edge API, allowing developers to deploy Gemini locally on supported Android devices, including their latest smartphone product line \cite{google24, semenova24, darra24, snap}. 
Microsoft markets their hybrid hosting concept called "PC + Copilot," in which a language model runs locally on laptops \cite{mehdi24}. 
LLM-based services on embedded devices are also rapidly gaining popularity. 
Locally deployed language models already power personal assistants on smartwatches, wearable speech-to-speech translators, and voice-controlled interfaces in smart home devices.
Adoption in various robotic systems is also growing. 
Industry-leading robotic products such as Tesla Optimus, Figure 02, and Unitree G1 all utilize a natural language user interface that enable more intuitive and efficient human-robot interactions.
Looking into the future of mobile operating systems, well-known proposals like LLM OS, advocate for LLM-based operating systems that interpret and execute commands based on natural language.
Recent works, including LLM as OS, AIOS, MemGPT, and open source projects, such as UI-TARS Desktop and phidata, all contribute to a similar concept of a natural-language-based computer interface. 
They collectively point to the same trend of even deeper integration of LLM on personal devices.

The strong interest in locally hosted LLMs is primarily motivated by the following factors.
First, (\textit{User Experience}) considering LLMs are becoming multi-modal, streaming multimedia data to the cloud induces significant latency into the system, negatively affecting the user experience. 
Second, (\textit{Connectivity Issues}) cloud-based inference services rely on a stable internet connection, a privilege that users don't always have on edge devices. 
Third, (\textit{Privacy Concerns}) streaming personal information to the cloud for Retrieval Augmented Generation (RAG) greatly increases the risk of privacy breaches \cite{lewis20}.  
Finally, (\textit{Compute Scalability}) the growing compute demand for AI services requires exponential growth in cloud compute capability. 
Locally hosted options that leverage edge devices' computing power could mitigate the impact of such demand and provide an alternative scaling option.
In summary, user experience, connectivity issues, privacy concerns, and compute scalability are the major driving forces for locally deployed LLMs.
Similar concerns have already motivated the local deployment of CNN onto various edge devices, including laptops, cell phones, cameras, etc.
It would not be a surprise to see LLM follow a similar trajectory. 

\subsection{Large Language Model Quantization}
Reducing the computational cost and memory consumption of LLMs is a pressing challenge, particularly when deploying them on edge devices. Quantization, a method used to approximate higher-precision data using lower-precision representations, has emerged as a standard solution to mitigating this issue. In this paper, we focus primarily on weight quantization methods that generally fall in one of two categories: quantization aware training (QAT) and post training quantization (PTQ). QAT methods simulate quantization during extensive retraining or finetuning phases to mitigate accuracy loss \cite{li23, nagel21}. Although these methods yield superior downstream accuracy, they are both time-consuming and incredibly taxing on computation and memory resources. As this burden is exacerbated with larger models, comparatively lightweight PTQ methods that require little to no training data are generally favored for quantizing LLMs \cite{Kim23, li23, lin24, park24, yao22}.

One such example is GPTQ, an exemplary one-shot, per-layer quantization method that iteratively quantizes parameters in blocks and uses approximate second-order information to update the remaining, yet-to-be-quantized weights, mitigating quantization loss \cite{frantar22}. Other effective PTQ methods include SpQR\cite{dettmers23}, Activation-Aware Weight Quantization (AWQ) \cite{lin24}, SmoothQuant \cite{xiao23}, ZeroQuant \cite{yao22}, AdaRound \cite{nagel20}, and SqueezeLLM \cite{Kim23}. In these methods, regardless of weight precision, actual computations usually rely on standard precision formats like as FP16 or FP8. Our elastic framework for hosting quantized LLMs on edge devices is incredibly versatile and can support any PTQ method to mix quantization levels as long as all computations use the same precision. In this paper, we mainly employ ExLlamaV2 for weight quantization. ExLlamaV2 is based on GPTQ and supports mixing quantization levels both across layers and within individual modules, making it particularly well suited for FlexQuant. ExLlamaV2 minimizes quantization error over a diverse calibration set while meeting a target average bitrate for the model. We find that this PTQ method offers the best trade-off between memory footprint, accuracy, and latency for our experiments.

\subsection{Elastic Serving}

Although PTQ methods are incredibly performant, it is not sufficient to serve only one static model when deploying models on edge devices where many different applications interact with the system, each with its own latency, memory footprint, and accuracy requirements. It would be impossible to meet these varying Service Level Objectives (SLOs) with a single model \cite{yin24, yin24a}. However, storing a large portfolio of models of varying sizes is also impractical. Practical deployment on edge devices demands an elastic solution that can dynamically serve different models in a cost-efficient way.

One state-of-the-art method for efficient deployment is Any-Precision LLM(AnyPrecision) \cite{park24}. Starting with a ``seed model'' quantized (using SqueezeLLM) to the minimum supported bit-width, Any-precision incrementally upscales the model by appending bits to the parameters of the model. At runtime, bits can be truncated or re-appended efficiently, and total storage costs are dictated solely by the largest precision. While this method is elastic, it does not provide a fine-grained transition in memory space as there are large jumps in model footprint between the supported precisions. Furthermore, the method does not identify how to dynamically set precision to meet different SLOs. FlexQuant's versatility allows us to deploy AnyPrecision within the FlexQuant framework, adding granularity to model transitions and providing an ideal trade-off curve between accuracy and memory-footprint.

\subsection{Transition Granularity}
\label{section:granularity}
We need to identify an acceptable upper bound on the size of the gaps in memory footprint between models served by the FlexQuant framework. These gaps should be small enough so that we can scale up or scale down the model size as memory is requested or freed by other programs on the edge device. 
It is equally important to minimize the memory IO overhead during swapping between different versions of the elastic model. 
At any point at runtime, we want to identify the model that fits in the current available memory, yet maximizes performance, leaving little "extra" memory, while minimizing memory transition overhead. 
If we want to swap out a block of parameters, we need to first load the replacement block of parameters in addition to the current hybrid model. We want the size of the additional block of parameters to be small.
 
One example scenario is deployment on mobile phones, where FlexQuant would ideally change model size as apps are opened and closed, suggesting that the ideal granularity is the size of an application. The authors of "End the Senseless Killing"\cite{EndSK} profile the memory allocated by eight popular apps on Android devices, finding that most use around 100MB or less. We additionally profiled application memory allocation over one month of typical usage on an Android phone. Of the top 30 most memory intensive apps in this time period, less than 10\% used more than 200MB, and the majority of applications used less than 100MB.
If we apply Anyprecison to this scenario, we are wasting, on average, over 400MB of memory. 
Thus, an efficient elasticity hosting method should provide at least 100MB of memory granularity and minimize memory IO overhead.

\section{FlexQuant Framework}
\label{section-methodology}
In this work, we propose a unified solution, FlexQuant, providing flexibility for one-shot deployment and dynamic memory flexibility at runtime.
For any input model, FlexQuant generates an ensemble of hundreds of quantized models with different memory footprints. 
The maximum difference in footprint between any two models is less than 100MB, which maximizes memory usage for the diverse deployment scenarios. 
At the same time, every model in the ensemble shares a compact pool of model parameters, as small as a pair of quantized models.  
The parameter sharing minimizes total storage footprint while ensuring minimal memory IO due to runtime model transition.

FlexQuant leverages the interchangeability of quantized parameters in different bit widths. 
Because an n-bit quantized LLM layer is a numerical approximation of its FP16 counterpart, representations of the layer in different bit-widths still share the same approximation target.
Thus, a quantized LLM does not experience a significant loss of accuracy if a module is replaced with a slightly lower bit-width counterpart. 
Empirical observation also supports this intuition.
Based on experiments on a Llama 2 7B model, the perplexity drop of an 8-bit model due to replacing one of its modules with a 3-bit version is always under $0.02$.
This characteristic suggests that a smooth memory-accuracy trade-off, through gradually replacing quantized LLM's layers with their lower bit-width counterparts, is attainable. 
FlexQuant leverages this characteristic to design a high-granularity ensemble with low storage cost. 

Before diving into the details of FlexQuant, we define some technical terms for the following discussion.
FlexQuant enables elastic hosting by generating an ensemble of Elastic Quantization Models ($EQM$). 
The models in the ensemble have gradually smaller memory footprints to enable memory elasticity.
The maximum difference in footprint between two adjacent models defines the ensemble's transition granularity. 
The total size of the entire ensemble determines its storage overhead. 
Existing ensemble design methods struggle to generate high-granularity memory transitions while maintaining low storage cost. 
FlexQuant automatically identifies $EQM$s to meet this need while improving the accuracy-memory trade-off. 
Its flexible mechanism also allows for adaptation to a wide range of quantization methods. 

\subsection{EQM Generation Philosophy}

Quantization compresses an FP16 model, $M$, into a quantized model, $QM(n)$, with an effective bit-width of $n$.
Given two effective bit-widths $n_{low}, n_{up}$ where $n_{low} < n_{up}$, FlexQuant generates a suite of $EQM$s by gradually replacing $QM(n_{up})$'s quantized layers with quantized layers from $QM(n_{low})$. 
$QM(n_{low})$ and $QM(n_{up})$ define the lower and upper bound of the $EQM$'s memory footprint, respectively.
Every $EQM(n_{low}, n_{up})$, defined by $QM(n_{low})$ and $QM(n_{up})$, only uses parameter from those two models and does not require extra storage cost. 
It could also incorporate additional $m$ quantized models $QM(n_{mid}^{i})$, with $n_{low} < n_{mid}^{i} < n_{up}$ and $0 \leq i < m$, to formulate a new set of $EQM(n_{low}, n_{mid}^{0}, ..., n_{mid}^{m-1}, n_{up})$, for better memory-accuracy trade-off. 

The benefits of such a design philosophy are two-fold. 
Firstly, the largest memory footprint difference between two elements in the $EQM(n_{low}, n_{up})$, is no larger than the size of a single LLM module, ensuring high-granularity.
Secondly, the high granularity does not require storing new parameters for every element in the $EQM(n_{low}, n_{up})$.
No matter how many module replacement trajectories we generate from $EQM(n_{low}, n_{up})$, they only utilize parameters from $QM(n_{low})$ and $QM(n_{up})$, which ensures low storage cost. 

\subsection{Efficient Design Space Navigation}
Although FlexQuant's $EQM$ generation method brings multifold benefits, it still suffers from a major technical challenge: navigating its massive design space. 
For an $EQM$ ensemble, formulated by $m+2$ QMs, every element in $EQM(n_{low}, n_{mid}^{0}, ..., n_{mid}^{m-1}, n_{up})$ has a candidate pool with the size of $(m+2)^{\#module}$, where $\#module$ indicates number of modules in a language model.
As a lower bound example, a 2 model $EQM$ ensemble for Llama-2 7B has a design space with the size of $2^{66} = 7.38*10^{19}$. 
To efficiently navigate a design space of this size, FlexQuant narrows down the available options by leveraging deployment constraints. 
In a realistic deployment scenario, in addition to transition granularity and storage cost, it is also important to limit the transitional memory cost due to module replacement. 
With this consideration, FlexQuant only allows for module transition from its current bit-with to its lower bit-width counterpart and forbids backward transition. 
This constraint minimizes the total memory IO cost during continuous transition.

With this design space constraint, the search for the next $EQM^k(n_{low}, n_{mid}^{0}, ..., n_{mid}^{m-1}, n_{up})$ simplifies into selecting which module to replace with its lower bit-width version. 
The upper bound of design space's complexity dramatically reduced to $(m+1)*\#module$. 
In the case of 2 model $EQM$ ensemble designed for Llama-2 7B, the complexity reduces to 66 for selecting the next configuration. 
However, this simplification is not enough. The total design space complexity of generating every configuration in the model ensemble, $EQM(n_{low}, n_{mid}^{0}, ..., n_{mid}^{m-1}, n_{up})$ could be as large as $  \prod_{i=1}^{(m+1)*\#module} i$, which is still too large. 

We use an approach that is inspired by Monte Carlo Tree Search\cite{coulom06}, and the search for $EQM$ generation could be viewed as a tree traversal process. 
The $EQM$ search starts from the $QM(n_{up})$, which could be viewed as the root of the tree structure. 
Stemming from the root, there are in total $(m+1)*\#module$ options as the next configuration.
All options would become their own subtree's root and the pattern goes on, until every stem reaches the end. 
All stems would end with the configuration reaching the $QM(n_{low})$ as leaf nodes. 
A complete tree traversal path from the root node to a lead node represents a valid $EQM$ suite. 

To efficiently navigate the tree, we follow a similar formulation to the Monte Carlo Tree search. 
FlexQuant only keeps $\#stem$ of the $EQM$ ensemble during search process. 
At the start of every iteration, FlexQuant loops through every last $EQM_{last}$ in all ensembles. 
Following the one-way design policy, every successor $EQM$, generated by replacing a module of those $EQM_{last}$ into a lower bit-width counterpart, is added into a candidate pool. 
To narrow down the scope, it utilizes a pre-generated sensitivity analysis to reserve only the top $\#branch$ candidate for every $EQM_{last}$. 
The sensitivity analysis is generated by replacing a single module from the $QM(n_{up})$ into its lower bit counterparts and calculating Euclidean distance between logit outputs from the hybrid model and $QM(n_{up})$.
Every module using a bit-width selected from ${n_{low}, n_{mid}^{0}, ..., n_{mid}^{m-1}}$, would be associated with a numerical value.
The value indicates the additional quantization error resulting from replacing such module with their $n_{up}$ counterpart.
The smaller the error, the higher its rank. 

After sensitivity analysis, FlexQuant directly evaluates model performance instead of relying on a pre-built sensitivity analysis.
To keep the search time reasonable, we utilize the same calibration set used in GPTQ and ExLlamaV2. 
The calibration set is a diverse collection of token sequences that generally reflect the active numerical range during actual deployment.  
It serves as a quick evaluation to estimate model performance without having to run a full-scale benchmark.
Its effectiveness has been proven in various previous works. 
FlexQuant's search method estimates the performance of each candidate $EQM$, before completing the entire tree traversal. 
Based on the evaluation results, top $\#stem$ $EQM$s are added into the ensemble and the search process moves onto the next iteration.
The iterative process will continue, until every $EQM_{last}$ in every ensemble has reached $QM(n_{low})$.
The ensemble with the best trade-off curve is selected as the final output.   

The pseudo code for FlexQuant's tree search algorithm is presented in \hyperref[alg:tree]{Algorithm~\ref*{alg:tree}}.

\begin{algorithm}
\caption{Tree Search for EQM Generation}
\label{alg:tree}
\begin{algorithmic}
\STATE $ensemble \gets List([QM(n_{up})])$
\STATE $EQM_{last} \gets GetLastEQM(ensemble)$
\WHILE {$EQM_{last} \neq QM(n_{low})$}
    \STATE $EQM_{cand} \gets CandidateGenerator(EQM_{last})$
    \STATE $EQM_{cand} \gets AnalysisFilter(EQM_{cand})$
    \STATE $ppl = list()$
    \FOR {$eqm \in EQM_{cand}$}
        \STATE $ppl.Append(CalibrationEval(eqm))$
    \ENDFOR
    \STATE $EQM_{cand}.DescendingRank(ppl)$
    \STATE $EQM_{new} \gets EQM_{cand}[0:\#stem-1]$
    \STATE $ensemble.Append(EQM_{new})$
    \STATE $EQM_{last} \gets GetLastEQM(ensemble)$
\ENDWHILE
\RETURN $ensemble$;
\end{algorithmic}
\end{algorithm}

\begin{figure}
    \centering
    \includegraphics[width=1.0\linewidth]{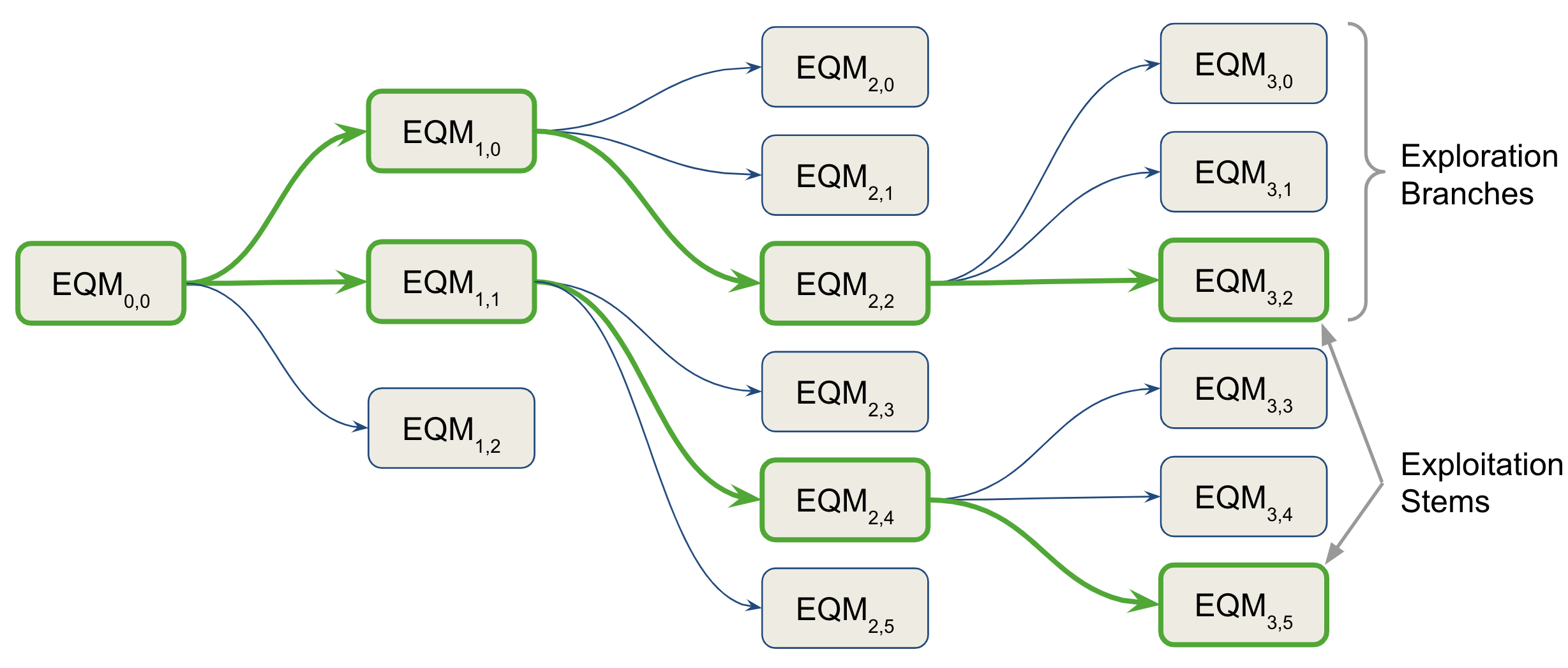}
    \caption{An example of FlexQuant's tree search for EQM ensemble. The shown search process has two exploitation stems and three exploration branches.}
    \label{fig:tree_search}
\end{figure}

\autoref{fig:tree_search} illustrates an example with two stems and three branches. 
At the start of the second iteration, the last $EQM$s from the ensembles are $EQM_{1,0}$ and $EQM_{1,1}$. 
All of their potential successor $EQM$s are added to the candidate pool. 
After going through the sensitivity analysis, only $EQM_{2,0}, EQM_{2,1}, ... EQM_{2,5}$ are selected for calibration set evaluation.
Only the top two stems, $EQM_{2,2}$ and $EQM_{2,5}$, are selected into the ensemble and become the new $EQM_{last}$ in the search process. 

\subsection{EQM Pruning Strategy}
As mentioned above, it is possible to generate an $EQM$ ensemble with more than 2 starting $QMs$ to improve its memory and accuracy trade-off. However, not all parameters from the $QM(n_{mid}^i)$ models are necessary nor do they contribute equally to improving the trade-off. Many parameters are not even used in the final $EQM$ ensemble. This observation inspired our pruning strategy that reduces the overhead introduced by $QM(n_{mid}^{i})$ models while retaining the accuracy improvements.

FlexQuant's pruning strategy relies on a ranking of the importance or effectiveness of every module introduced by $QM(n_{mid}^{i})$.
Each module's rank is determined by the number of models in the $EQM$ that leverage its parameters.  
For modules that have similar amount of usage, ranks are determined by whether the $EQM$ utilizes them at lower footprints.
As shown in \autoref{figure:intro}, the accuracy of an $EQM$ is much more sensitive to changes in parameters at smaller memory footprints, shown by the decreasing gradient of the perplexity trade-off curve.
Thus, all modules from $QM(n_{mid}^{i})$ are ranked first by their overall usage in $EQM$ ensemble and then fine-tuned by their usage for models at lower footprints. 
Based on this ranking system, FlexQuant can prune the generated $EQM$ ensemble, providing additional storage savings. Additionally, the ranking system allows FlexQuant to prune the ensemble at different rates, granting the user more flexibility in balancing accuracy and storage costs.

\begin{algorithm}

\caption{EQM Pruning Stratigy}
\label{alg:prune}
\begin{algorithmic}
\STATE $module\_dict \gets Dict()$
\STATE $qm \gets List([QM(n_{mid}^{0}), QM(n_{mid}^{1}), ..., QM(n_{mid}^{m-1})])$
\STATE $ensemble \gets TreeSearch(qm)$
\FOR {$model \in qm$}
    \FOR {$module \in model$}
        \STATE $module\_dict[module] \gets 0$
    \ENDFOR
\ENDFOR
\FOR {$eqm \in EQM_{cand}$}
    \FOR {$module \in eqm$}
        \STATE $module\_dict[module] ++$
    \ENDFOR
\ENDFOR
\STATE $modules \gets module\_dict.Keys()$
\STATE $counts \gets module\_dict.Values()$
\STATE $modules.DescendingRank(counts)$
\STATE $modules.DescendingRank(counts)$
\STATE $\#modules \gets Len(modules)$
\STATE $modules \gets modules[0:prune\_rate*\#modules-1]$
\RETURN $modules$;
\end{algorithmic}
\end{algorithm}

\section{Experiment}
\label{section-experiment}

We implement our method using ExLlama on Llama 1 7B, Llama 2 7B, and Llama 3 8B and using AnyPrecision on Llama 1 7B and Llama 2 7B \cite{touvron23a, touvron23b, dubey24}. For both settings, we generate our $EQM$ ensemble using Exllama's built-in calibration set containing a mixture of data including C4, wikitext, code, and multilingual data. 
We evaluate this $EQM$ ensemble containing a continuum of hybrid models by measuring both their memory footprint and downstream accuracy. 
We measure the hybrid model's perplexity on C4 \cite{Raffel20}, WikiText2 \cite{merity16}, and PTB \cite{marcus93}.
In addition to measuring perplexity, we evaluate downstream accuracy on ARC Easy, ARC Challenge \cite{clark18}, Hellaswag \cite{zellers19}, PIQA \cite{bisk20}, and WinoGrande \cite{sakaguchi21}.
These benchmarks cover a diverse range of tasks and domains of applications, which demonstrates our $EQM$s' performances in real deployment scenarios.

\begin{figure*}[ht]
\centering
\includegraphics[width=1.0\textwidth]
{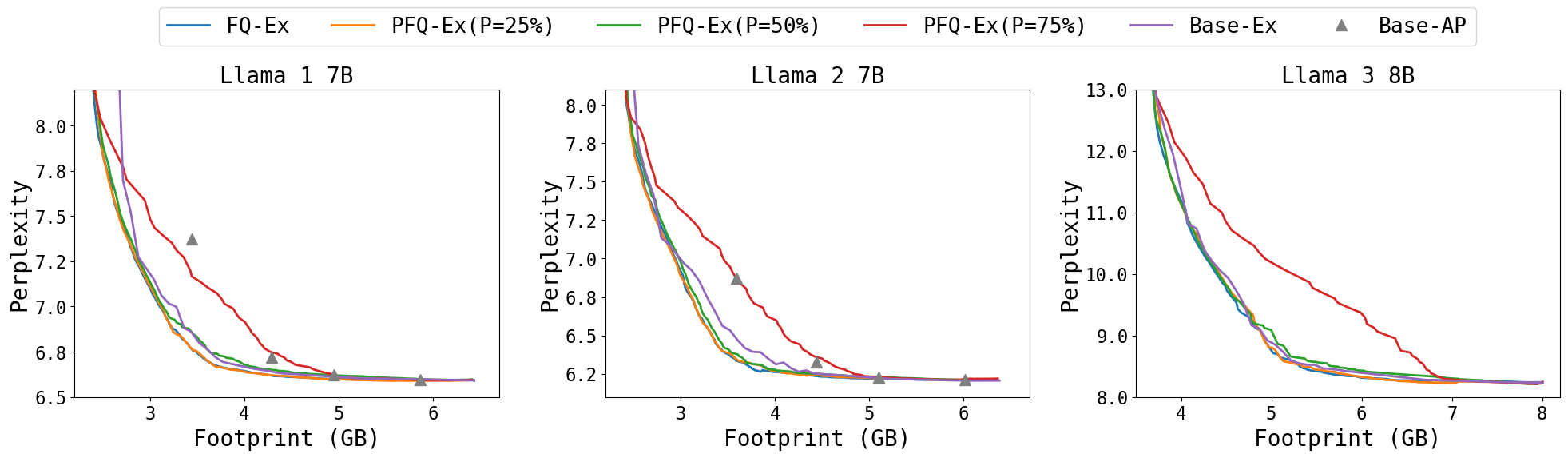}
\caption{Perplexity comparison between Base-Ex, Base-AP, FQ-Ex, and PFQ-Ex at different pruning rate. Results for AnyPrecision on Llama 3 8B is omitted due to lack of support in their implementation. The footprint range is different due to differences in parameter count and model architecture.}
\label{figure:pruning}
\end{figure*}
\begin{figure*}[ht]
    \centering
    \includegraphics[width=1.0\textwidth]
    {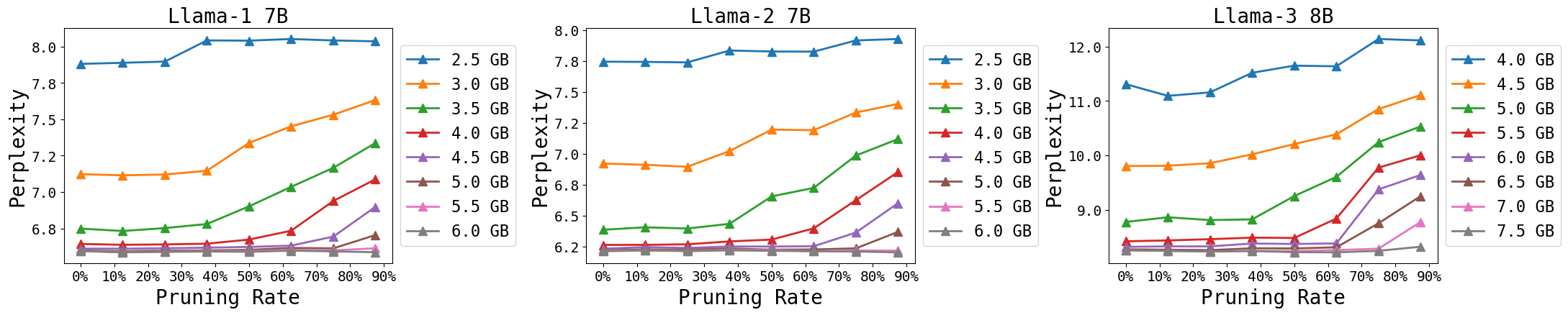}
    \caption{Perplexity vs pruning rate at varying memory footprint bounds for FQ-Ex on Llama 1, Llama 2 and Llama 3}
    \label{figure:pruning_comparison}
\end{figure*}

\section{Results \& Discussion}
\label{section-results}

\subsection{Calibration Set Accuracy Results}
\autoref{figure:intro} shows how FlexQuant, when applied to ExLlamaV2 (FQ-Ex) and to AnyPrecision (FQ-AP), compares to baseline elastic hosting framework using ExLlamaV2 (Base-Ex) and AnyPrecision (Base-AP). We compare these methods by measuring perplexity on the calibration set using Llama 2 7B. Both FQ-Ex and FQ-AP yield smooth transition trajectories with a granularity of $\sim$100MB. In other words, every consecutive point along each FlexQuant curve represents a potential hybrid model that is only around 100MB larger than the previous one. In comparison, replicating this elastic transition trajectory at 100MB granularity using Base-Ex requires a model ensemble using 177GB in storage, over 10x that of FQ-Ex. Replicating this 100MB granularity with Base-AP is not possible at all. Not only does FQ-Ex require an order of magnitude less storage, it also provides the best memory footprint-accuracy trade-off. \autoref{figure:intro} shows that FQ-Ex is almost always at the Pareto frontier.

\subsection{Pruning}
\begin{figure*}[ht]
\centering
\includegraphics[width=0.95\textwidth]
{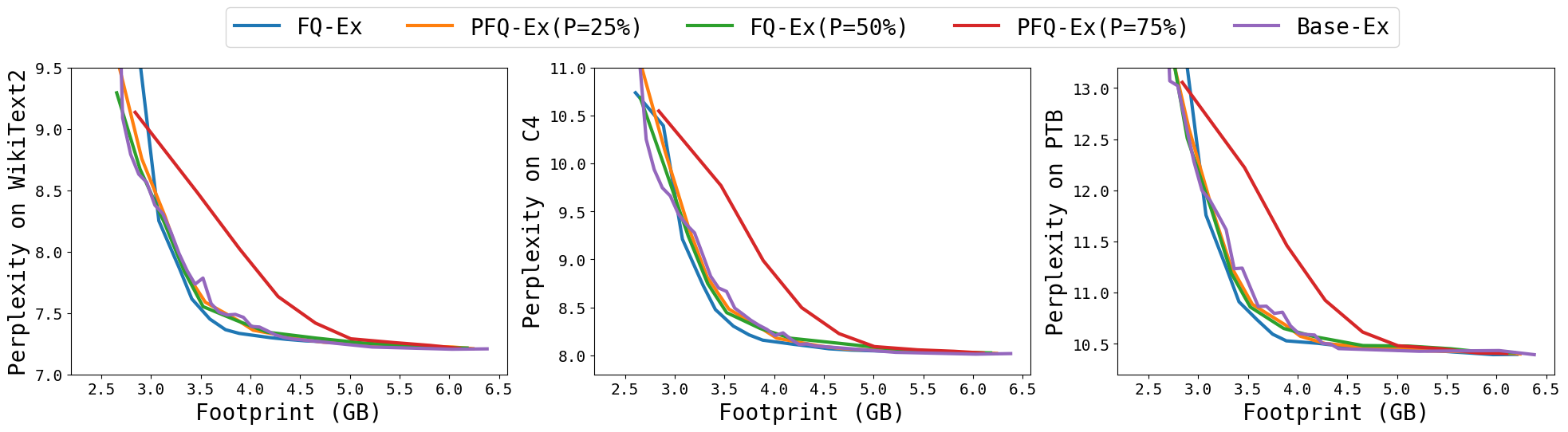}
\caption{Downstream perplexity comparison of quantized Llama models between Base-Ex FQ-Ex and PFQ-Ex at different pruning rate.}
\label{figure:downstream_ppl}
\end{figure*}

\begin{table*}[htbp]
\caption{Downstream task accuracy comparison between Base-Ex, FQ-Ex and PFQ-Ex with 50\% pruning.}
\def\arraystretch{1.5}
\tabcolsep=0.17cm
\fontsize{7pt}{7pt}\selectfont
\begin{center}
\begin{tabular}{|c|c|c|c|c|c|c|c|c|c|c|c|c|c|c|c|c|}
\hline
  \multicolumn{2}{|c|}{Task} & \multicolumn{3}{c|}{$arc_{Challenge}$} & \multicolumn{3}{c|}{$arc_{Easy}$} & \multicolumn{3}{c|}{$hellaswag$} & \multicolumn{3}{c|}{$piqa$} & \multicolumn{3}{c|}{$winogrande$} \\
\hline
 \multicolumn{2}{|c|}{Policy} & Base-Ex & FQ-Ex & PFQ-Ex & Base-Ex & FQ-Ex & PFQ-Ex & Base-Ex & FQ-Ex & PFQ-Ex & Base-Ex & FQ-Ex & PFQ-Ex & Base-Ex & FQ-Ex & PFQ-Ex \\
 \hline
 \multicolumn{2}{|c|}{Storage} & 177 GB & 17 GB & 13 GB & 177 GB & 17 GB & 13 GB & 177 GB & 17 GB & 13 GB & 177 GB & 17 GB & 13 GB & 177 GB & 17 GB & 13 GB \\
 \hline
 \hline
 \multirow{7}{*}{\rotatebox[origin=c]{90}{Memory Footprint (GB)}}
 & 3.0 & 37.40 & -1.40 & -1.40 & 68.98 & +1.73 & +1.73 & 51.47 & -0.60 & -0.80 & 75.63 & +0.15 & +0.54 & 66.61 & +0.24 & -1.10 \\
 \cline{2-17}
 & 3.5 & 39.42 & +2.04 & +1.08 & 72.05 & +2.57 & +1.77 & 53.21 & +3.02 & +1.87 & 76.22 & +1.14 & +0.60 & 66.69 & +2.22 & +1.34 \\
 \cline{2-17}
 & 4.0 & 41.38 & +1.19 & +1.11 & 75.01 & +0.35 & +0.20 & 55.81 & +0.76 & +0.55 & 77.04 & +0.76 & +0.71 & 68.27 & +1.10 & +0.40 \\
 \cline{2-17}
 & 4.5 & 43.17 & -0.26 & -0.43 & 75.46 & +0.00 & -0.25 & 55.81 & +0.93 & +0.77 & 77.75 & +0.26 & +0.00 & 69.37 & +0.32 & -0.39 \\
 \cline{2-17}
 & 5.0 & 43.17 & +0.17 & -0.25 & 75.46 & +0.13 & -0.17 & 56.85 & +0.08 & +0.03 & 78.01 & +0.11 & +0.12 & 69.38 & +0.39 & -0.16 \\
 \cline{2-17}
 & 5.5 & 43.34 & +0.17 & -0.26 & 75.58 & +0.51 & +0.01 & 56.92 & +0.20 & +0.07 & 78.12 & +0.07 & +0.06 & 69.38 & +0.37 & +0.23 \\
 \cline{2-17}
 & 6.0 & 43.34 & +0.34 & +0.08 & 75.58 & +0.51 & +0.01 & 57.03 & +0.14 & -0.01 & 78.18 & +0.05 & +0.11 & 69.45 & +0.44 & +0.16 \\
 \hline
 \multicolumn{17}{l}{\textit{NOTE: Results of FQ-Ex and PFQ-Ex are presented as accuracy differences, comparing to their Base-Ex counterpart at the same storage footprint.}}
\end{tabular}
\end{center}
\label{tab:downstream}
\end{table*}
As FQ-Ex offers the best trade-off between memory footprint and accuracy, we investigate whether this method can be further improved using pruning. \autoref{figure:pruning} compares Base-Ex, Base-AP, FQ-Ex, and FQ-Ex with pruning (PFQ-Ex).
PFQ-Ex is implemented with four pruning rates, P=25\%, 50\%, and 75\%. The four PFQ-Ex curves were all obtained with an $QMs$ of 2.5, 4.0, 5.0, 7.5 bits while varying pruning rates between 0, 0.25, 0.5, ad 0.75, yielding storage costs of 17GB, 15GB, 13GB and 11GB respectively. 

Compared to PFQ-Ex, Base-Ex requires over 10x more storage, and Base-AP is roughly 15x less granular. We repeat this experiment using Llama 1 7B, Llama 2 7B, and Llama 3 8B. The memory footprint ranges we explore differ between these three models because Llama 3 8B has more parameters than the other two models. Similarly, the perplexity ranges we achieve differ mainly due to different starting performance for the base models. We find that FQ-Ex, PFQ-Ex with P=25\%, and 50\% all consistently match or outperform Base-Ex and Base-AP for all three models. 

Our pruning method adds more flexibility to FlexQuant. A user can use the same set of quantized models to generate an entire family of trade-off curves with different storage limits by simply adjusting the pruning rate. Starting with the same set of quantized models, FlexQuant can identify multiple candidate hybrid models with similar memory footprints by varying the pruning rate.
That is, hybrid models of two different precisions may satisfy the same upper bound depending on how we adjust the pruning rate. \autoref{figure:pruning_comparison} demonstrates that models with approximately the same memory footprint can yield different calibration set perplexity depending on the pruning rate at search. The results suggest that relatively low pruning rates maintain similar level of performance as FQ-Ex. However, performance takes a significant hit at around P=40\%.

\subsection{Downstream Task Performance}
Measuring perplexity on the calibration set alone is not sufficient to confirm model performance. 
It is important that FlexQuant performs well on downstream tasks that are representative of queries encountered in real-world on-device deployment. We evaluate our method's performance with Llama 2 7B, as shown in \autoref{figure:downstream_ppl} and \autoref{tab:downstream}. 

\autoref{figure:downstream_ppl} compares perplexity on downstream/non-calibration datasets (WikiText2, C4, and PTB) between the ExLlamaV2 baseline and FlexQuant(ExLlamaV2) implemented at four different pruning rates, p=0\%, 25\%, 50\%, and 75\%. The top three FlexQuant configurations match or outperform the ExllamaV2 baseline for all tasks. We again see that pruning reduces footprint, but a large pruning rate can destabilize the transition curve and greatly impact performance. \autoref{figure:downstream_ppl} demonstrates a dramatic decrease in accuracy when increase pruning rate from 50\% to 75\%. The observations from \autoref{figure:pruning_comparison} and \autoref{figure:downstream_ppl} suggest that it is not advisable to apply pruning rates greater than 40--50\%.

\autoref{tab:downstream} further compares accuracy on downstream tasks. At 0.5GB intervals, we compare downstream accuracy of the models at that specific memory footprint interval yielded by Base-Ex, FQ-Ex and PFQ-Ex with a 50\% pruning rate. For example, the ``3.0 GB'' row details the downstream task accuracy of the FQ-Ex and PFQ-Ex hybrid models at 3.0 GB compared to the downstream task accuracy of the Base-Ex model directly quantized to 3.0 GB. As shown in the table, the baseline requires 177GB in storage while FQ-Ex and PFQ-Ex only consume 17GB and 13GB of storage respectively.
Despite this greater than ten-fold decrease in storage costs, FQ-Ex and PFQ-Ex hybrid models at a given precision typically not only match baseline performance, but also sometimes surpass the downstream task accuracy given by the Base model. 
This increase in performance is due to our tree search process.
The search process is essentially an additional refinement on the quantized LLM, which further optimizes its performance under a footprint limit. 
As a conclusion, even though our search process is conducted on the calibration dataset, we are presumably not overfitting on this calibration set and sacrificing downstream performance compared to the ExLlamaV2 quantized models.

\begin{figure*}[ht]
\centering
\includegraphics[width=0.95\textwidth]
{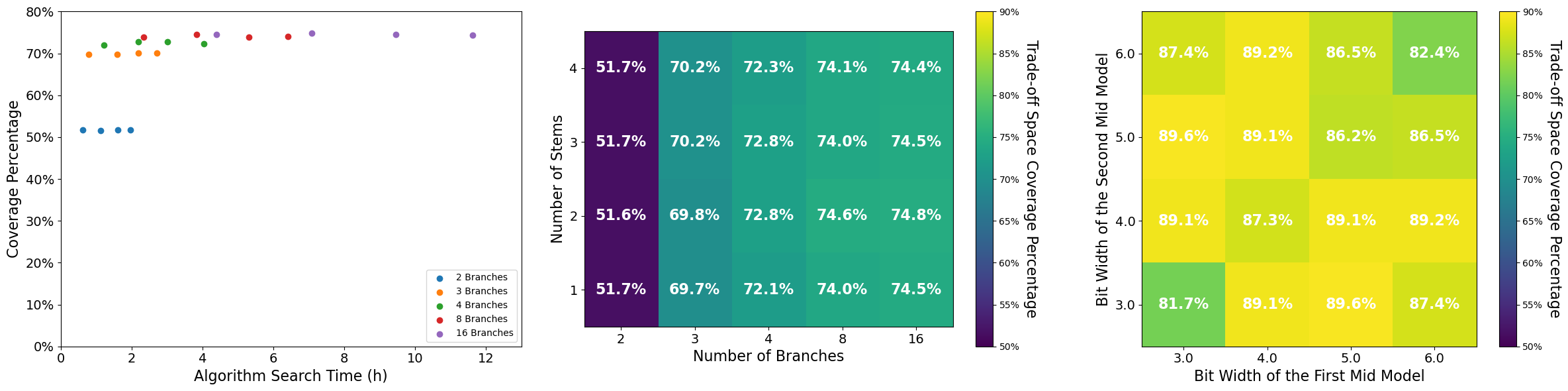}
\caption{Hyperparameters' impact on the overall search algorithm's performance. In the left and middle plots, we sweep between different numbers of stems and the number of branches. The right plot shows the impact of the middle model. In the middle and right plots, the lighter the color is, the better the quality of the search result.}
\label{figure:coverage_percentage}
\end{figure*}
\begin{figure*}[ht]
\centering
\includegraphics[width=0.95\textwidth]
{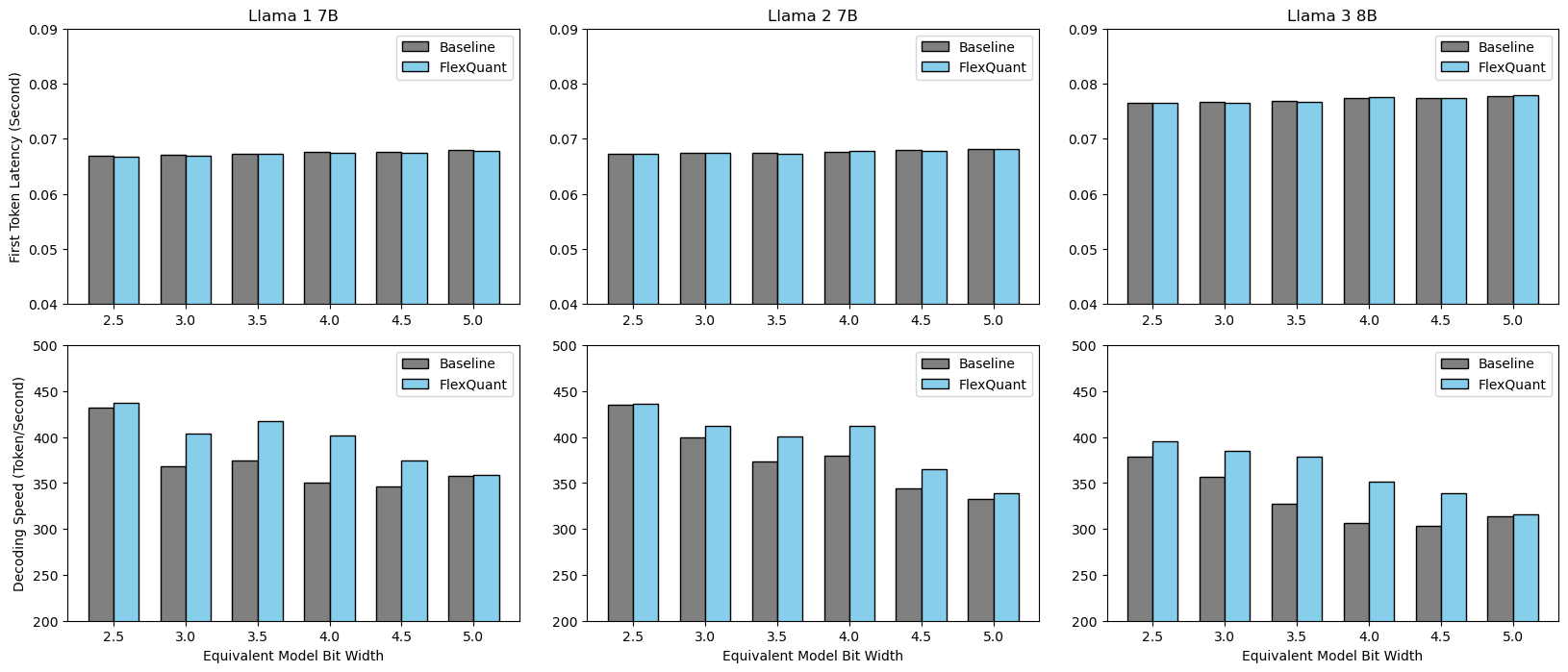}
\caption{First token latency and decoding throughput overhead of FlexQuant framework. The Baseline is Base-Ex, indicated by the grey bar. The FlexQuant is FQ-Ex, indicated by the light blue bar.}
\label{figure:framework_overhead}
\end{figure*}

\subsection{Ablation Study}
The number of stems and the number of branches are two of the key hyperparameters during the search process. 
We conduct a series of ablation studies to evaluate their impact on the overall search time and quality of the search result. 
To unify our standard for the quality of the search results, we formulate an evaluation metric, called "Coverage Percentage".
In the perplexity and footprint trade-off space, we define the ideal Coverage Area (ICA), as the rectangular area defined by the smallest model and the largest model.
The Coverage Area (CA) is the area enclosed by the trade-off curve, generated from a FlexQuant tree search. 
The coverage percentage is defined as the ratio of the ICA to the CA of a search result. 
The higher the percentage is, the better the search result. 

The left and the middle plots in \autoref{figure:coverage_percentage} demonstrate the comparison between the number of branches and the number of stems. 
The compassion is based on a search process with only the largest and the smallest models. 
The middle plot is color-coded according to the coverage percentage of the search results. 
It clearly shows that the number of branches makes a much greater impact on the overall search strategy than the number of stems. 
The improvement saturates at around 16 branches. 
The left plot also shows a similar trend. 
As more branches are included, the overall search time increases, but the coverage percentage remains roughly unchanged. 
Imperially, the added number of stems improves search process stability, especially for search processes with more than two models. 
The right plot demonstrates the search results, using one or two middle models during the search process. 
All color blocks with the same x-axis and y-axis values represent the usage of one middle model, while all other blocks represent search processes with two middle models.
Compared to the middle plot, even adding one additional model could dramatically improve the overall search quality. 
Adding two models would further improve the quality and reach a coverage percentage as high as 89.6\%. 

\subsection{Framework Overhead}
During inference, each layer of EQM only contains parameters from one QM. 
FlexQuant loads in new EQMs layer-by-layer during swapping instead of all layers together. 
Thus, the memory footprint overhead during swapping never exceeds the size of one layer, less than 2\% of the total size of an EQM. 
As for memory IO overhead, say 300MB of memory footprint becomes available, based on benchmarking, the total memory IO cost averages around 324MB(11\% overhead), and never exceeds 395MB (24\% overhead). 
Admittedly, some swapping processes could be costly (>1GB memory IO cost).
However, they only occur after a significant change in available memory, which happens infrequently. 
FlexQuant does not introduce additional computational overhead. 
The smallest model shows up to an 18\% improvement in decoding speed.
Other EQMs between the smallest and the largest EQM show decoding speed within that 18\% range. 
To further validate our claims, we compare our FlexQuant performance with the ExLlamaV2 baseline across all Llama models. 
Our results are shown in the \autoref{figure:framework_overhead}.
We cut off the equivalent model bit widths at 2.5, since a lower bit width would result in the model being unusable.
A higher than 5.0 is usually unnecessary, as a FlexQuant @ 5bits would perform very similarly to its FP16 counterparts.
Sweeping across different equivalent model bit widths, FlexQuant's first token latency is identical to the baseline, without any noticeable overhead.
As for the decoding speed, FlexQuant not only shows no overhead but also demonstrates better performance over the baseline. 

\section{Case Studies}
To demonstrate the effectiveness of FlexQuant in an actual edge system, we also evaluate FlexQuant based on traces collected from a commercial smartphone, Samsung S23 Ultra, with the latest Android 14 operating system. 
We execute six popular applications (YouTube, Samsung Internet, Google Maps, TikTok, X(Twitter), Subway Surf) to collect realistic mobile workload traces.
We use Perfetto, an opend-source tool by Google for system tracing and benchmarking on Android, Chrome OS or Linux, to collect memory usage traces. 
For every trace, we collect the following metrics: MEMINFO\_MEM\_TOTAL, MEMINFO\_MEM\_FREE, MEMINFO\_MEM\_AVAILABLE, MEMINFO\_BUFFERS and MEMINFO\_CACHED.
To simulate realistic user behavior, we apply a similar methodology as proposed in Ariadne \cite{Ariadne}. 
We use monkeyrunner to generate a series of user instructions in each app. 
Emulating user behavior with monkeyrunner eliminates inconsistency due to human error and ensures our results are reproducible. 
Three collections of workload traces are collected, based on the number of applications concurrently running in the system.  
Those collections present a type of workload with 2, 4, and 6 active applications, respectively. 
For every collection, we randomly generate a user command sequence, including launching an application and various user commands within the application. 
The emulation length of each sequence is no less than 10 minutes.
We generate 100 different sequences of user inputs and repeat measurement of each sequence five times to minimize random variation.
Once all traces have been collected, we aggregate them into those three collections.
The distribution of memory usage, reflected in those traces, is shown in \autoref{figure:mem_usage}.
In the following sections, we utilize those three collections of workload traces to demonstrate the effectiveness of FlexQuant in various runtime scenarios. 

\subsection{LLM Performance Comparison}
\begin{figure*}[ht]
\centering
\includegraphics[width=0.95\textwidth]
{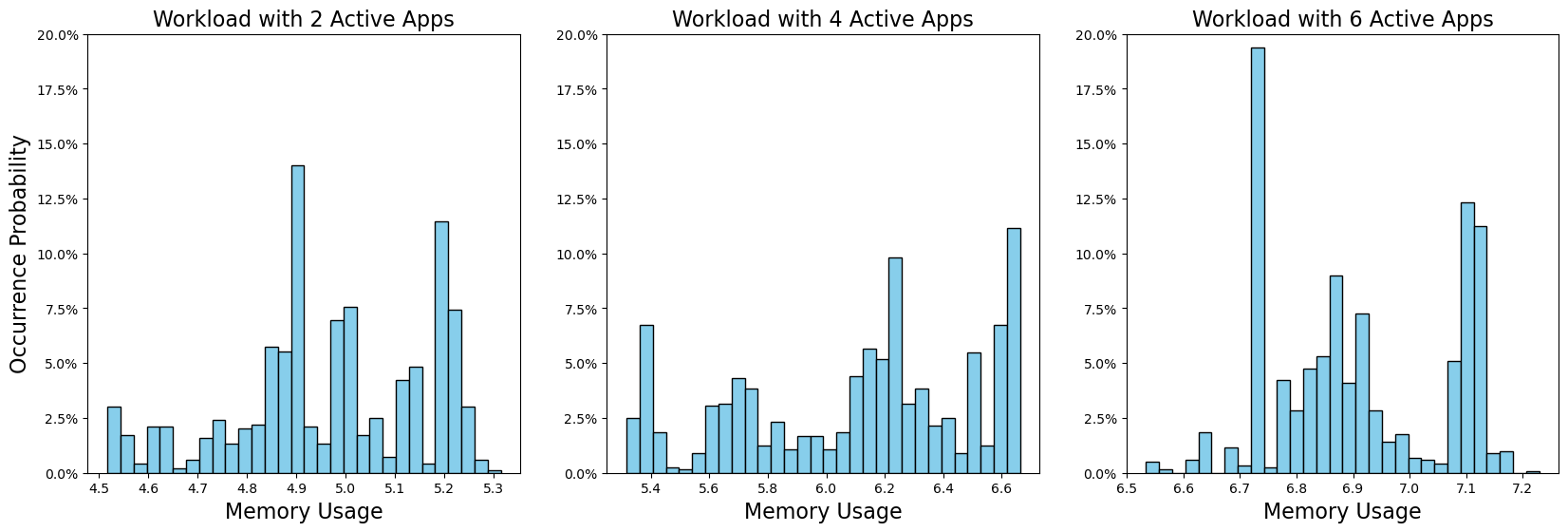}
\caption{Memory usage collected on a Samsung S23 Ultra, using the Android 14 operating system. Three collections of workload traces are collected based on the number of active apps on the cellphone. Their memory usage histograms are presented from left to right.}
\label{figure:mem_usage}
\end{figure*}
\begin{figure*}[ht]
\centering
\includegraphics[width=0.95\textwidth]
{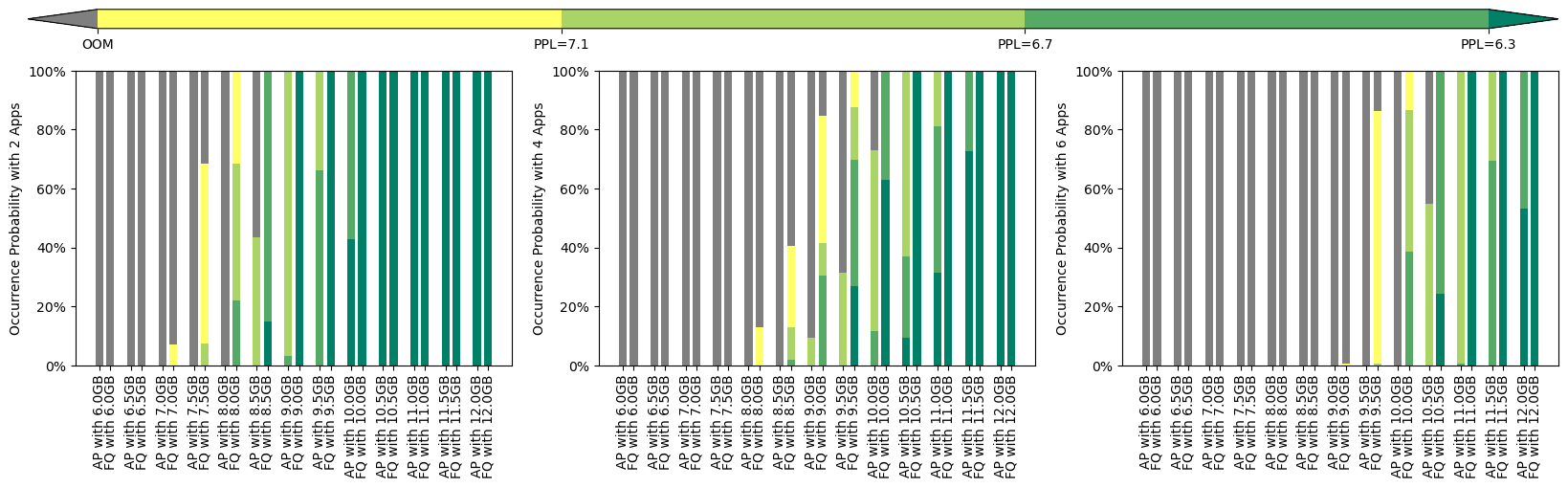}
\caption{Model quality comparison under different workload scenarios. "AP with 6.0GB" or "FQ with 6.0GB"  means AnyPrecision or FlexQuant deployed on a device with 6GB of total system memory. The lighter the color is, the better the model quality.}
\label{figure:system_ppl}
\end{figure*} 
Considering most LLMs deployed on edge devices serve as AI assistants, these LLMs could be launched at any instance during runtime. 
Our collected cellphone workload traces provide a realistic collection of memory usage on the cellphone.
Assuming that at the time of the launch, the LLM could only use the available compute memory in the system, we compare the launched LLM's performance in perplexity. 
The higher the chance of launching a model with better performance, the better the framework is. 
We sweep across different total system memories under those three collections of system workloads. 

\autoref{figure:mem_usage} visualizes our results. 
Each plot uses traces collected with 2, 4, or 6 active applications.
Within each plot, we compare AnyPrecision(AP) with FlexQuant(FQ) under different total system memory. 
The colored bar visualizes the performance of the launched LLM under the memory constraint.
As indicated by the top color legend, the darker the color, the better the performance. 
Grey color suggests that even the smallest model fails to meet the memory constraint.
Due to the variance of available memory, the LLM's performance with the same workload may vary. 
Take "FQ with 10.0GB" with 4 active applications as an example. 
Around 60\% of the time, the launched LLM's PPL is lower than 6.3, while in the rest of the time, its PPL is higher than 6.3 and not greater than 7.1. 
Comparing those colored bars side by side, the colored bar with more coverage of a darker color indicates the better solution.
Regardless of total system memory or workload traces, FlexQuant consistently outperforms AnyPrecision in the performance of LLMs. 

\subsection{LLM Launch Speed Comparison}
\begin{figure*}[ht]
\centering
\includegraphics[width=0.95\textwidth]
{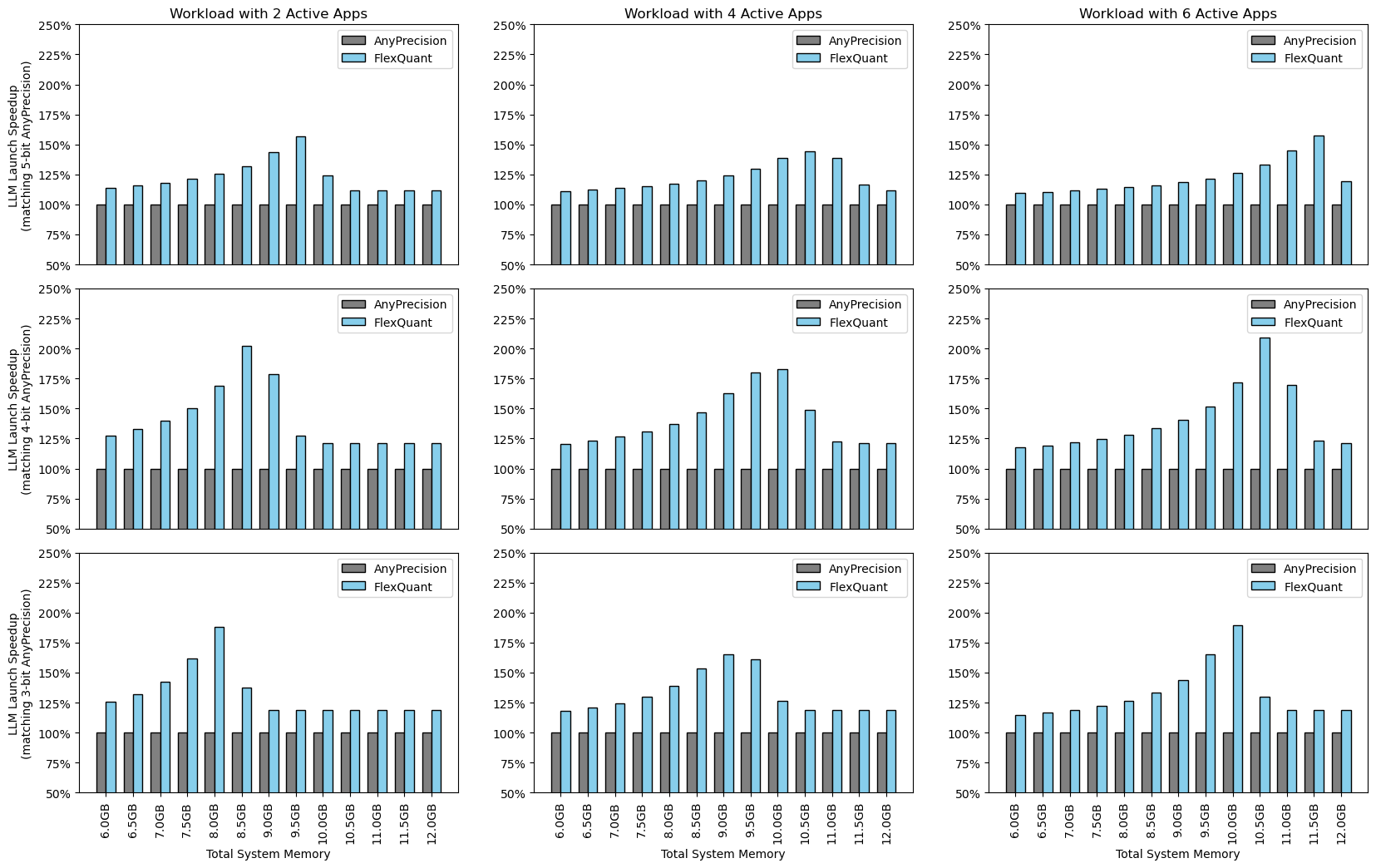}
\caption{LLM launch speed comparison between FlexQuant (FP-Ex) and AnyPrecision (Base-AP). The larger the speedup is, the better the performance.}
\label{figure:launch_speed}
\end{figure*}
Apart from launching LLM with strict memory constraints, another common runtime scenario is launching with memory swapping. 
Memory swapping, also known as virtual memory or paging, is a memory management technique used by mobile operating systems to extend the apparent amount of available RAM by utilizing storage space (flash memory) to hold less frequently used data. This allows devices to run more applications concurrently and handle memory-intensive tasks, but can also come with performance implications. 
When physical RAM is nearly full, the OS detects the need for more memory.
The OS identifies data in RAM that is not currently actively used, such as background apps or inactive processes.
The identified data is moved from RAM to a dedicated area of storage called the swap space (swap partition or swap file).
The vacated RAM is now available for other applications or processes.
When the swapped-out data is needed again, it is retrieved from storage and brought back into RAM, potentially displacing other less active data in RAM. 
It should be noted that memory swapping is only needed when the LLM can not fit into available system memory. 
If the LLM could already fit into the memory, there is no need for memory swapping at all.

With memory swapping, an LLM could always launch with the same level of performance regardless of the available memory at the time of launch. 
However, it comes with an overhead of additional latency due to memory swapping.
Thus, a better framework should achieve a better launch speed, given the same level of LLM performance. 
To compare FlexQuant with AnyPrecision, we set three LLM performance standards, based on AnyPrecision's 3, 4, and 5 bits models. 
For every scenario, we ensure that the AnyPrecision and the FlexQuant models have the same level of performance based on these three standards. 
We sweep across different workload traces and performance standards. 
The comparison is presented in \autoref{figure:launch_speed}. 
It is evident that FlexQuant outperforms AnyPrecision across the board. 
Interestingly, all the plots follow a similar trend with FlexQuant's speedup peaking in the middle. 
Those peaks result from FlexQuant launching the LLM without the need for memory swapping, while AnyPrecision still requires memory swapping.
If there is more total system memory, the speedup is less significant and eventually reaches a plateau when both AnyPreicsion and FlexQuant no longer need memory swapping.
With less total system memory, FlexQuant also requires memory swapping, which reduces its speedup over AnyPrecision.

\section{Conclusion}
\label{section-conclusion}

Growing reliance on LLMs, coupled with the ever-present need for personalization and privacy, has led to increasing interest in locally hosted LLMs. 
One of the major challenges with locally hosting LLMs is memory elasticity. 
Existing elastic serving approaches either cannot dynamically adjust model sizes in small increments or fail to scale due to significant storage cost. 
Our FlexQuant work demonstrates a viable direction to greatly improve transition granularity and storage cost. 
Future system designers could leverage FlexQuant to have more flexibility and design options when deploying LLMs on edge devices.
Additionally, we believe that a LLM's usage of other key hardware resources, such as energy or compute, also need to be adjusted elastically, a direction we aspire to explore in future works. 

\bibliographystyle{IEEEtran}
\bibliography{8-references}

\vfill

\end{document}